\pdfoutput=1

\documentclass[11pt]{article}

\usepackage[final]{acl}

\usepackage{times}
\usepackage{latexsym}
\usepackage{amsfonts}
\usepackage{amsmath}
\usepackage{subcaption}
\usepackage{amssymb}
\usepackage{todonotes}

\usepackage{titlesec}

\usepackage[T1]{fontenc}

\usepackage[utf8]{inputenc}

\usepackage{microtype}

\usepackage{inconsolata}

\usepackage{graphicx}
\usepackage{soul}
\usepackage{booktabs}

\title{A framework for analyzing concept representations in neural models}

\author{
 \textbf{Burin Naowarat},
 \textbf{Hao Tang},
 \textbf{Sharon Goldwater}
\\
 The Centre for Speech Technology Research \\
 School of Informatics, University of Edinburgh, United Kingdom
\\
 \small{
   \href{mailto:burin.naowarat@ed.ac.uk}{burin.naowarat@ed.ac.uk},
   \href{mailto:hao.tang@ed.ac.uk}{hao.tang@ed.ac.uk},
   \href{mailto:sgwater@inf.ed.ac.uk}{sgwater@inf.ed.ac.uk}
 }
}

\begin{document}

\maketitle

\begin{abstract}
Understanding how neural models represent human-interpretable concepts is challenging. Prior work has explored linear concept subspaces from diverse  perspectives, such as probing and concept erasure. We introduce a unified framework to study these subspaces along two axes: \textit{containment}, which tests if a concept is fully represented in a subspace but not outside it, and \textit{disentanglement}, which tests for isolation from other concepts. 
In experiments on both text and speech models,
we first highlight that concept subspaces may not be uniquely determined, and discuss the implications for concept subspace analysis. Then, we compare properties of concept subspaces estimated using five estimators, proposed in different communities. We find that (1) the choice of estimator impacts the containment and disentanglement properties;  (2) the state-of-the-art concept erasure method, LEACE, performs well on both testing axes, but still struggles to generalize to unseen data; and (3) in HuBERT speech representations, phone information is both contained and disentangled from speaker information, while speaker information is hard to contain in a compact subspace, despite being disentangled from phones.\footnote{We release the source code at \url{https://github.com/burin-n/concept_space}.}

\end{abstract}

\begin{figure}[t]
  \centering
  \includegraphics[width=1\linewidth]{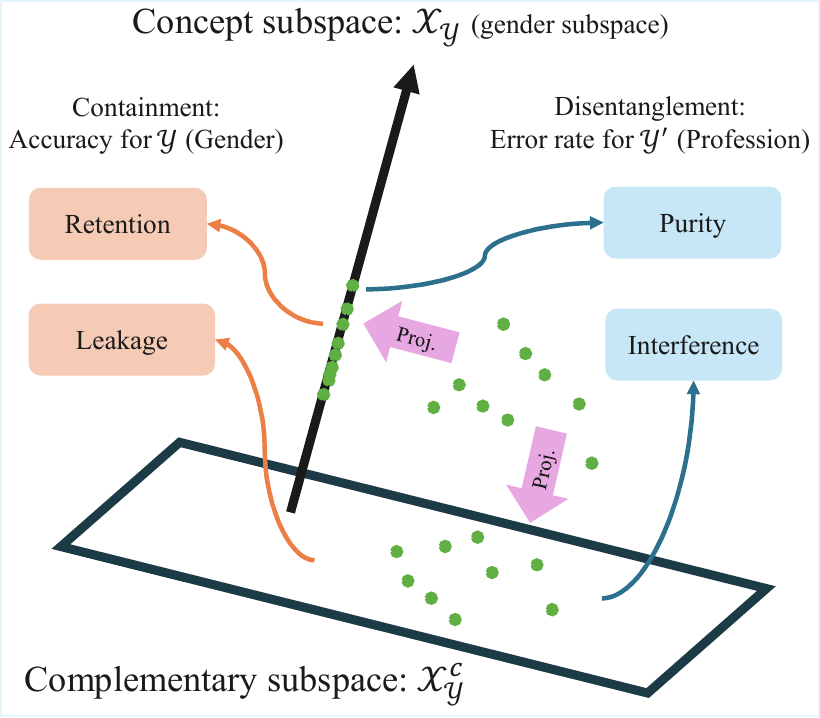} 
  \caption{An example concept subspace w.r.t.\ two concepts $\mathcal{Y}$ and $\mathcal{Y}'$. In this example, $\mathcal{Y}$ is for gender classes and $\mathcal{Y}'$ is for profession. Gender accuracies on $\mathcal{X}_\mathcal{Y}$ and $\mathcal{X}_\mathcal{Y}^c$ quantify retention and leakage, indicating how well gender information is contained in $\mathcal{X}_\mathcal{Y}$ (containment metrics, in orange). Profession error rates on $\mathcal{X}_\mathcal{Y}$ and $\mathcal{X}_\mathcal{Y}^c$ quantify purity and interference, indicating how well profession information is disentangled from $\mathcal{X}_\mathcal{Y}$ (disentanglement metrics, in blue).}
  \label{fig:framework}
  \vspace{-10pt}
\end{figure}

\section{Introduction} 

A long-standing question in cognitive science and AI is how \emph{internal representations} relate to \emph{behaviorally relevant} or \emph{human-interpretable} concepts. 
Empirical studies have suggested that in deep models of language, many binary distinctions and relations (such as past vs.\ present tense, or "capital-of") are represented as linear directions in the representation space \cite{mikolov_linguistic_2013,bolukbasi_man_2016,Park2023-xq,huben_sparse_2024}, and that more generally, multi-class concepts (such as part-of-speech) are captured in low-dimensional subspaces \cite{Tenney2019-ad, Coenen2019-bn, Yang2021-jy, Chang2022-ky, Hernandez2021-nv,huang_ravel_2024}. 
Despite known exceptions \cite{Csordas2024-dn} and limitations \cite{Ravfogel2022-zl}, linear models underlie much work in interpreting and controlling deep models of speech and language.

Nevertheless,
different studies have (often implicitly) made different assumptions about what it means to say a concept is represented in a subspace, and thus how we would evaluate that claim. For example, word analogy and probing studies \cite[e.g.][]{mikolov_linguistic_2013,belinkov_analyzing_2017,Tenney2019-ad,pasad_comparative_2023} have focused on whether concepts (or the relations defined by them) are \emph{decodable} from a subspace, where that subspace is implicitly defined by finding the directions that best differentiate the concept labels or exemplars. Meanwhile, concept erasure studies \cite{Ravfogel2020-jj,Ravfogel2022-cg,Belrose2023-ny} have focused on whether concepts are \emph{disentangled}: they aim to explicitly
identify a concept subspace that can be removed while retaining as much information as possible about other concepts. As well as focusing on different goals, these and other studies have used a range of different methods to identify the subspace(s) of interest, but with little or no systematic comparison between these methods.

In this paper, we formalize a framework for characterizing 
how concepts are represented, both individually and in relation to each other. This framework clarifies the assumptions behind different lines of work on concept representation, highlights the role of the concept (subspace) estimation method, and leads to four measures that can evaluate different kinds of claims. 

More specifically, our measures  capture two higher-level properties: the degree to which a particular subspace \textbf{contains} all information about a concept, and the  degree to which the concept subspace is \textbf{disentangled} from other concepts of interest.
In particular, given a concept and a corresponding subspace, we measure containment via \textbf{retention} (how much information about the concept can be decoded from the linear subspace) and \textbf{leakage} (how much information about the concept can be identified \emph{outside} of the linear subspace). 
We measure disentanglement via \textbf{purity} (how much information about \emph{other} concepts is encoded in the linear subspace) and \textbf{interference} 
(how much information about \emph{other} concepts is affected by linear manipulation of the concept subspace).
Related measures have been proposed for causal interventions on LLMs \cite{huang_ravel_2024}, but our measures are more widely applicable, and we illustrate them on speech and masked language models.

Our experiments on text (BERT) and speech (HuBERT) models illustrate how the proposed framework can shed light on the way concepts are represented and show that conclusions can be affected by the choice of concept estimation method. 
First, we use gender and profession concepts in BERT to demonstrate that the four measures are complementary and to emphasize that concept subspaces may not be uniquely determined, which has implications for structural analyses of concept representations. 
Next, we conduct a comprehensive comparison of concept estimators for phone and speaker concepts in HuBERT. The results indicate that (1) LEACE, an estimator from the concept erasure literature \cite{Belrose2023-ny}, finds concept subspaces with strong overall properties but exhibits leakage on unseen data, despite its no-leakage guarantee on seen data; and (2) while many estimators are able to find contained and disentangled phone subspaces, all fail to find speaker subspaces with good containment when testing with unseen speakers. 
Lastly, we further examine this speaker containment limitation and show that estimating a speaker subspace that is well-contained, compact, and capable of generalizing to unseen speakers is non-trivial.
Overall, our experiments demonstrate the importance of the proposed measures for concept subspace analysis and illustrate how the framework can be used to draw novel insights into concept representations.

\section{Concept representations}
\label{sec:method:concept_space}

We begin with our definition of a concept and our framework for exploring how concepts are represented. We use $\mathcal{X}$ to denote a $D$-dimensional representation space (for example, of model activations).
We define a concept $\mathcal{Y}$ as the set of possible outcomes of a random variable.
In text representations, for example, gender and profession are both concepts with a small discrete set of labels,\footnote{Both linguistic gender and social gender are concepts in text; here we refer to social gender, and define it as a binary concept so we can compare to prior work \cite{Belrose2023-ny}.}
while in speech representations, phonetic classes and speaker identity are also discrete concepts. 
Though alternative definitions of concepts are possible \cite{Park2023-xq, Belrose2023-ny, Kim2018-vm, dalvi2022discovering}, our definition is consistent with considerable prior work, including studies that probe for concepts \cite[e.g.][]{belinkov_analyzing_2017,Tenney2019-ad},  examine their geometric structure \cite{Chang2022-ky,Hernandez2021-nv}, seek to remove them from the representation space \cite{Ravfogel2020-jj,Ravfogel2022-cg}, or intervene to change one outcome to another \cite{huang_ravel_2024}.

We define a concept subspace for the concept $\mathcal{Y}$ as a subspace $\mathcal{X}_{\mathcal{Y}} \subseteq \mathcal{X}$.
The fact that $\mathcal{X}_{\mathcal{Y}}$ is a subspace means that it can be described as a span of basis elements $y_1, \dots, y_K \in \mathcal{X}$, or
\begin{align}
\label{eq:method:colspace}
\mathcal{X}_{\mathcal{Y}} = \text{span}(\{y_1, \dots, y_K\}).
\end{align}
The goal of estimating a concept subspace becomes finding a set of basis elements that spans  that space, and we discuss several estimators in Section~\ref{sec:method:space_estimation}.

\subsection{Containment and disentanglement measures}

According to these definitions, any subspace $\mathcal{X}_{\mathcal{Y}}$ of $\mathcal{X}$ can be considered a concept subspace.
This raises the question: what makes a \emph{good} concept subspace?
The answer may depend on the application or research question, but several criteria seem relevant.
First, information about $\mathcal{Y}$ that is represented in $\mathcal{X}$ should be \textbf{well-contained} in $\mathcal{X}_{\mathcal{Y}}$: that is, represented in $\mathcal{X}_{\mathcal{Y}}$ but not outside it.
We operationalize this idea via probing classifiers  \cite{Belinkov2022-pq, Conneau2018-wy}, defining \textbf{retention} as the accuracy of a classifier for $\mathcal{Y}$ when trained on $\mathcal{X}_{\mathcal{Y}}$, and \textbf{leakage} as the accuracy when trained on its complement $\mathcal{X}_\mathcal{Y}^c = \mathcal{X} \setminus \mathcal{X}_{\mathcal{Y}}$. These measures are illustrated in Fig.~\ref{fig:framework}.\footnote{Within our framework, non-linear probing classifiers could in principle be used to explore whether concepts are represented in low-dimensional linear subspaces, even if the concept labels are not linearly separable. In this paper, we illustrate the framework using linear probes.}
Note that if $\mathcal{X}$ contains redundant information about $\mathcal{Y}$, then (as we will see) retention and leakage can both be high, weakening any arguments based on the assumption that $\mathcal{X}_\mathcal{Y}$ is ``the'' subspace representing $\mathcal{Y}$. In contrast, strong containment is indicated by high retention and low leakage.

While containment focuses on how a concept is represented individually, \textbf{disentanglement} provides a complementary view focusing on its relationship to other concepts. 
Here, we draw inspiration from \citet{Higgins2018-kc} and reinterpret their disentanglement definition through the lens of linear concept erasure \cite{Belrose2023-ny}. The ideal disentanglement occurs when a concept subspace represents \textit{only one} concept,
and when erasing that concept subspace will not damage the representation of other concepts.
For simplicity, we consider pairs of concepts, so suppose we have another concept $\mathcal{Y}'$.
We define the \textbf{purity} of the concept subspace $\mathcal{X}_\mathcal{Y}$ as how poorly $\mathcal{Y}'$ is represented in $\mathcal{X}_{\mathcal{Y}}$, operationalized as the classifier error rate of predicting $\mathcal{Y}'$ on $\mathcal{X}_{\mathcal{Y}}$.
We define \textbf{interference} as how much damage happened to the representation of $\mathcal{Y}'$ after projecting to $\mathcal{X}_\mathcal{Y}^c$ (erasing $\mathcal{Y}$), i.e., the error rate of predicting $\mathcal{Y}'$ on $\mathcal{X}_\mathcal{Y}^c$.
A concept subspace is disentangled from another concept if it has high purity and low interference.\footnote{In the concept erasure literature \cite{Ravfogel2020-jj, Belrose2023-ny}, our leakage and interference measures are referred to as \emph{guardedness} and \emph{collateral damage}; retention and purity are not typically measured.}

Table~\ref{tab:method:metrics} summarizes the proposed metrics with their theoretical best- and worst-case bounds. We establish these boundaries from the principle that, under linear transformations,
the subspace being evaluated, $\mathcal{X}_\mathcal{Y} \subseteq \mathcal{X}$, cannot contain more decodable information about $\mathcal{Y}$ than the ambient space $\mathcal{X}$. For retention, the best case is that $\mathcal{X}_\mathcal{Y}$ is as informative as $\mathcal{X}$ in representing $\mathcal{Y}$. Conversely, the worst case occurs when $\mathcal{X}_\mathcal{Y}$ does not represent $\mathcal{Y}$ at all, meaning that predicting $\mathcal{Y}$ using features from $\mathcal{X}_\mathcal{Y}$ is no better than a majority-class baseline \cite{Xu2020-uw, Ravfogel2022-cg, Belrose2023-ny}. The same applies to leakage, purity, and interference.\footnote{In practice, care must be taken when comparing the scores with their respective best and worst cases. In particular, the containment is with respect to the best retention on the ambient space, and is less meaningful when the retention on the ambient space is itself low. Similarly, good disentanglement (low interference and high purity) is not possible to attain if the concepts themselves are fundamentally entangled.}

\begin{table}
    \scalebox{0.75}{
    \begin{tabular}{l|l|r|r}
    \toprule
         Metric & Measure & Best case& Worst case\\
         \midrule
          Ret. ($\uparrow$) & $\mathcal{Y}$ acc. on $\mathcal{X}_\mathcal{Y}$&$\mathcal{Y}$ acc. on $\mathcal{X}$& Maj.acc. for $\mathcal{Y}$\\
          Leak. ($\downarrow$) & $\mathcal{Y}$ acc. on $\mathcal{X}_\mathcal{Y}^c$& Maj.acc. for $\mathcal{Y}$& $\mathcal{Y}$ acc. on $\mathcal{X}$\\
          Pur. ($\uparrow$)  & $\mathcal{Y}'$ err. on $\mathcal{X}_\mathcal{Y}$& Maj.err. for $\mathcal{Y}'$& $\mathcal{Y}'$ err. on $\mathcal{X}$\\
          Int. ($\downarrow$) & $\mathcal{Y}'$ err. on $\mathcal{X}_\mathcal{Y}^c$&$\mathcal{Y}'$ err. on $\mathcal{X}$& Maj.err. for $\mathcal{Y}'$\\
        \bottomrule
    \end{tabular}
    }
    \caption{A summary of evaluation metrics for the concept subspace $\mathcal{X}_\mathcal{Y}$ w.r.t.\ two concepts $\mathcal{Y}$ and $\mathcal{Y}'$.  \textit{Ret.}, \textit{Leak.}, \textit{Pur.}, and \textit{Int.} refer to retention, leakage, purity, and interference. We denote  accuracy and error rate as \textit{acc.} and \textit{err.}, and \textit{Maj.}~is the majority-class predictor. 
    }
    \vspace{-5pt}
    \label{tab:method:metrics}
\end{table}

\subsection{Relationship to other proposals}

The assumptions underlying our formalization and measures differ from some other related proposals.
In machine learning, for example,  disentanglement measures assume that the true generative factors are known, and that each factor ideally corresponds to a single dimension of the representation space \cite{Carbonneau2022-hh, Eastwood2022-la}. These assumptions do not hold for complex human-interpretable concepts learned from natural data.
In NLP, \citet{huang_ravel_2024} proposed a causal framework for measuring disentanglement of concept subspaces in  LLMs. However, it relies on properties of LLM generation that do not hold for masked language models or speech models, our focus here.
In speech, \citet{mohebbi_disentangling_2024} proposed an information theoretic framework to \emph{supervise} disentangled representation learning. Here, we focus on subspaces estimated \emph{without explicit disentanglement supervision}, thereby preserving the original structure of the representation space.
In addition, \citet{Mohamed2024-xe} proposed a measure of orthogonality between concept spaces which is conceptually similar to our interference measure, but is based on variance rather than probing accuracy.
We discuss their work further in Section~\ref{sec:sm_exp}.

\section{Estimating concept subspaces}
\label{sec:method:space_estimation}

Estimating a concept subspace through finding its basis is not always straightforward.
It is sometimes easier to estimate a linear projection $P$ that maps $\mathcal{X}$ to $\mathcal{X}_\mathcal{Y}$, where by definition of linear projection, $P^2 = P$.
Expressing a concept subspace $\mathcal{X}_{\mathcal{Y}}$ as an image of a projection $P$ is also convenient, because its complement $\mathcal{X}_{\mathcal{Y}}^c$ is the kernel of $P$ (or the image of $I - P$).
If we manage to obtain a matrix $Y$ whose columns form the basis of a concept subspace, we can take the projection $P = UU^\top$, where $U$ is the matrix in $Y = U \Sigma V^\top$ after compact singular value decomposition (SVD).

\subsection{Subspace estimators}
\label{sec:estimators}

We consider five supervised concept estimators, all of which estimate either a matrix $Y$ whose column space is the concept subspace, or a projection $P$ whose image is the concept subspace. 
Most of these methods have been used as concept estimators before (see Appendix~\ref{appx:estimation} for example studies), but have not been directly compared. 
We do not consider unsupervised methods such as PCA or Sparse Autoencoders \cite{huben_sparse_2024}, which may find interpretable concept subspaces post hoc, but cannot be used to analyze given concepts of interest. We also do not consider causal supervised methods designed for LLMs
\cite[e.g.,][]{geiger_finding_2024,huang_ravel_2024}, since 
it is unclear how to apply them to the types of models we examine here. 
Our estimators are:

\titlespacing*{\paragraph}{0pt}{5pt}{3pt}

\paragraph{Multinomial logistic regression (MLR)} is just the linear classifier $p(y | x) \propto \exp(W x + b)$ where $W$ is the weight matrix and $b$ is the bias for $x \in \mathcal{X}$ and $y \in \mathcal{Y}$.
It finds directions that best separate concept classes.
Here the matrix $Y$ that spans the concept subspace is simply the weight matrix $W^\top \in \mathbb{R}^{D \times |\mathcal{Y}|}$.

\paragraph{Linear discriminant analysis (LDA)} is a linear classifier that additionally assumes the classes are Gaussian with the same covariance matrix. LDA finds a subspace that maximizes the ratio of between-class variance to within-class variance. The matrix $Y$ consists of the eigenvectors of $\Sigma^{-1} \Sigma_b$, where $\Sigma$ is the within-class covariance matrix and $\Sigma_b$ is the between-class covariance matrix.

\paragraph{Centroid principal component analysis (CPCA)} \cite{Liu2023-sq} arranges the centroids (i.e., means) of the classes into a matrix $X \in \mathbb{R}^{D \times |\mathcal{Y}|}$, and the concept subspace is the span of the principal components of $X$.
Given that there is a subsequent SVD, we can simply take $Y = X$.

\paragraph{Covariance of input representation and concepts (COV)}
has been used as an approach to concept erasure \cite{Shao2023-nn}. Here, we compute $\Sigma \in \mathbb{R}^{D \times |\mathcal{Y}|}$, the covariance matrix between the input representation and the concept (represented as one-hot vectors), and take $Y = \Sigma$.

\paragraph{LEACE} \cite{Belrose2023-ny} is
another concept erasure method, which takes one step beyond COV and uses whitening.
We still compute the covariance matrix $\Sigma$ between the input representation and the concept one-hot vectors,
but the LEACE-estimated subspace is the image of the projection $P=W^{-1} U U^\top W$, where $W \in \mathbb{R}^{D \times D}$ is the whitening matrix and $U$ consists of the left-singular vectors of the whiten covariance matrix $W\Sigma$ after compact SVD.
To erase a concept, LEACE applies the projection $I - P$.
Note that $P$ is generally an oblique (non-orthogonal) projection.

\paragraph{Random projection (RAND)}
is a simple baseline where the matrix $Y$ consists of random Gaussian vectors sampled from $\mathcal{N}(\mathbf{0}, I/|\mathcal{Y}|)$.

\titlespacing*{\paragraph}{0pt}{5pt}{8pt}

\subsection{Generalization of concept subspaces}

To test the generalization of concept subspaces, we partition the dataset into three disjoint subsets: \textit{space-train}, \textit{probe-train}, and \textit{test} sets. We estimate concept subspaces using the space-train set, train the probing classifier using the probe-train set, and report the scores on the test set.
Prior work on concept erasure often uses the same sets for estimating concept subspaces and training probing classifiers  \cite{Ravfogel2022-cg, Belrose2023-ny}, and under this setup, LEACE is guaranteed to have zero leakage.
In contrast, we use three disjoint sets to ensure that the evaluation reflects the generalization of both the concept subspaces and the classifiers.

\section{The estimator matters: gender and profession subspaces in BERT}
\label{sec:lm_exp}

We first study gender and profession subspaces in text representations using two concept estimators, MLR and LEACE.
The goal is to see our framework in action and to show that different subspace estimators can lead to drastically different containment and disentanglement measures.
This finding highlights the danger of drawing strong conclusions about how concepts are represented based on a single estimator.

We use the same task and data as \citet{Belrose2023-ny}, who investigated LEACE as a way to mitigate gender bias in a profession classifier by erasing the gender concept from BERT representations. In addition to the leakage and interference measures reported by \citet{Belrose2023-ny}, we also report retention and purity. We do the same for MLR, which is widely used for probing but has also been used to analyze the geometry of concept representations \cite{Hernandez2021-nv}.

We follow \citet{Belrose2023-ny}, conducting our experiments on the Short Biography dataset \citep{De-Arteaga2019-mv}.
The dataset is labeled with 28 professions and 2 genders, and the concepts in this case are profession and gender.
Every paragraph in the dataset is embedded with BERT into a single vector (the vector that corresponds to the CLS token).
We use the training set to identify the profession and gender subspaces, the development set to train the probing classifier, and the test set to evaluate containment and disentanglement.

\subsection{Concept containment}
\label{sec:lm_exp:containment}

\begin{figure}[t]
\centering
\begin{subfigure}{0.7\linewidth}
    \includegraphics[width=\linewidth]{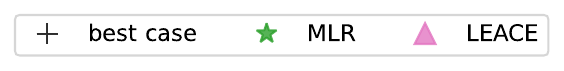}  
\end{subfigure} \\
\vspace{-4pt}
\begin{subfigure}{.49\linewidth}
    \includegraphics[width=\linewidth]{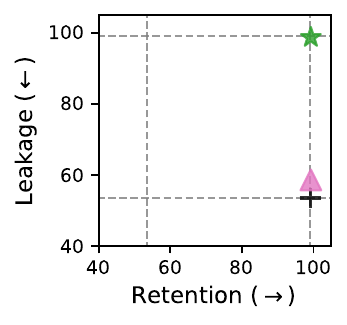} 
  \caption{Gender containment}
  \label{fig:lm_exp:main:sfig1}
\end{subfigure}
\begin{subfigure}{.49\linewidth}
  \includegraphics[width=\linewidth]{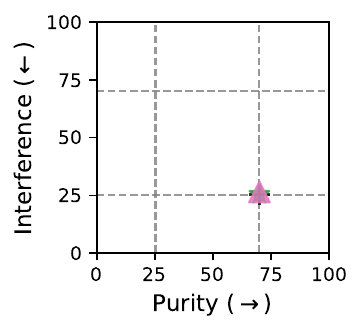} 
  \caption{Gender disentanglement}
  \label{fig:lm_exp:main:sfig2}
\end{subfigure}
\begin{subfigure}{.49\linewidth}
    \includegraphics[width=\linewidth]{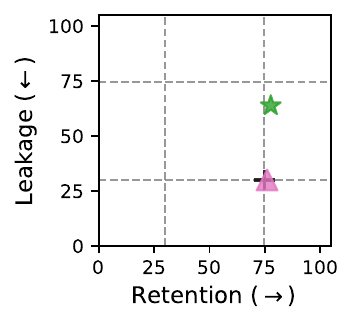} 
  \caption{Prof. containment}
  \label{fig:lm_exp:main:sfig3}
\end{subfigure}
\begin{subfigure}{.49\linewidth}
  \includegraphics[width=\linewidth]{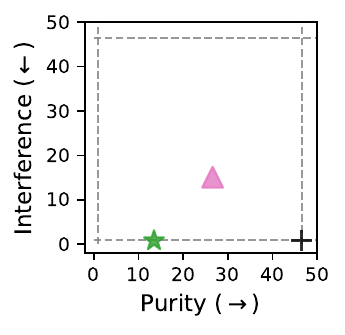} 
  \caption{Prof. disentanglement}
  \label{fig:lm_exp:main:sfig4}
\end{subfigure}
\caption{
Containment and disentanglement for gender and profession (Prof.) subspaces in BERT representations. Subspaces are estimated using MLR or LEACE. 
For each metric, the dashed lines show best- and worst-case boundaries (as described in  Table~\ref{tab:method:metrics}), and the arrows displayed on each label indicate directions towards the best cases.
For numerical results and results from other estimators, see Appendix \ref{appx:exp:lm}, Table~\ref{tab:lm_exp:main_frozen_space} and Figure~\ref{fig:full_text_exp:main_fz}.
}
\label{fig:lm_exp:main}
\vspace{-1em}
\end{figure}

We first focus on the containment properties of the gender and profession concepts, shown in the left column of Fig.~\ref{fig:lm_exp:main}. 
In terms of \emph{retention}, both estimators are
optimal, matching the result on the ambient space (right-hand dashed lines).\footnote{In fact, profession retention (Fig.~\ref{fig:lm_exp:main:sfig3}) slightly exceeds that of the ambient space, as probes on low-dimensional subspaces generalize better to the test set.}
This indicates that both estimators can find subspaces that preserve concept information well.
This result is not surprising for MLR, which is optimized for high retention, but has not previously been reported for LEACE, which is optimized for low leakage.

In contrast to retention, we observe a sharp divergence in \emph{leakage} between the two estimators: LEACE (as expected) has near-optimal leakage for both gender and profession, but MLR displays the opposite behavior, suggesting that substantial concept information is encoded outside the estimated concept subspace.

While leakage in MLR-estimated subspaces is known (in fact, it motivated the development of LEACE), we believe this is the first work to show that MLR and LEACE find distinct high-retention subspaces.
This implies that high-retention subspaces may not be unique---that is, there can be redundancy in the representation space.

The implications go beyond these two estimators, since previous work on interpreting the structure of concept subspaces has typically used retention-based estimators \cite{Hernandez2021-nv, Chang2022-ky, Park2024-ox}. However, if there are multiple high-retention subspaces for a concept, then conclusions about the structure of that concept (alone or in relation to other concepts) may not hold for all of these subspaces.
We argue that it is important to consider both retention, which measures concept information \emph{inside} the concept subspace, and leakage, which measures concept information \emph{outside} the concept subspace, to determine whether a concept is well-contained in a subspace; and only if so is it reasonable to draw further conclusions about the concept representation based on that subspace.

\subsection{Concept disentanglement}
\label{sec:lm_exp:disentanglement}

Next, we discuss disentanglement, shown in the right column of Fig.~\ref{fig:lm_exp:main}. 
As with containment, we see that different estimators can find subspaces with different disentanglement scores (see Fig.~\ref{fig:lm_exp:main:sfig4}). However, disentanglement has some additional properties arising from the fact that it measures relationships between distinct concepts. In particular, it can depend on the number of concept classes (dimensionality of the estimated subspace), and is not symmetric---that is, just because $\mathcal{Y}$ is disentangled from $\mathcal{Y}'$ does not imply the opposite.

To illustrate these points, we first consider the gender concept, which has only two classes. Fig.~\ref{fig:lm_exp:main:sfig2} shows that both estimators find subspaces with optimal disentanglement, despite differences in containment. In particular, erasing the gender subspace does not interfere with profession (low interference). Although higher interference could be possible, low interference is not surprising since a single-dimensional subspace is unlikely to overlap with directions encoding other concepts (especially if there is redundancy in the representation space). On the other hand, since the gender subspace has only a single dimension, it is unlikely for a multi-class concept to be decodeable from it (high purity).

Second, we turn to the profession concept, which has 28 classes. Although the estimators have different scores, both have sub-optimal purity, meaning that gender is partly decodable from their estimated profession spaces. In addition, the LEACE-estimated subspace shows noticeable interference (i.e., erasing it makes gender less decodable from the complement space). This indicates that the LEACE-estimated profession and gender subspaces overlap, even though this overlap is not detected by the disentanglement measures of the gender subspace, due to its low dimensionality. This finding further highlights the asymmetry in these measures and the role of concept dimensionality. Meanwhile the lack of interference in MLR, despite low purity, is another indication of redundant information in the representation space. Finally, we note that the disentanglement properties cannot be predicted based on the containment measures, demonstrating their complementary roles.

\section{Analyzing phone and speaker subspaces}
\label{sec:sm_exp}

In the previous section, we showed the utility of our framework, revealing new insights in the estimated concept subspaces and their properties.
We now apply our framework to the speech domain, where disentanglement has been studied in prior work \cite{Liu2023-sq, Mohamed2024-xe, gub25} in terms of two concepts that  convey different types of information:
phonetic categories and speaker identity.
The goal is to use our framework to complement prior work and to give a more detailed picture of how phonetic and speaker subspaces interact.

The phonetic categories, which encode linguistic content, are ideally speaker invariant. This can be achieved if the concepts are disentangled.
Empirically, prior findings indeed indicate that phone and speaker subspaces, estimated via CPCA, are approximately orthogonal in many self-supervised speech models \cite{Liu2023-sq, Mohamed2024-xe, gub25}, implying disentanglement. However, these prior studies have limitations: 
they did not measure containment or purity of the concept spaces (probing only the full representation space), 
and only reported unidirectional interference (speakers on phones), as well as their own variance-based disentanglement measure. 
In addition, the quality of the CPCA estimator itself remains untested relative to other estimators.

To address these gaps, we conduct a more comprehensive evaluation of phone and speaker subspaces, comparing all estimators in Section~\ref{sec:estimators}.
We studied representations from HuBERT \cite{Hsu2021-qo}, a widely-used Transformer-based self-supervised speech model.
We focused on the 11$^{\text{th}}$ HuBERT layer, which is best for phonetic tasks \cite{Chen2021-fe}.
We extracted frame-level features for LibriSpeech dev-clean and test-clean sets \cite{Panayotov2015-xr}, and used Kaldi forced alignment
to obtain frame-level phone labels.
The 5-hour dev-clean set was used as the space-train set. For each speaker in the test-clean set, we kept 70\% of their spoken frames (240 mins) for probe-train and used the remaining 30\% of frames (100 mins) as the test set. While the identical 39 phoneme classes were shared across the three data subsets, the speaker identities were disjoint. 
The space-train set had unique 40 speakers (the same number used by \citet{Liu2023-sq, Mohamed2024-xe}), 
while probe-train and test sets shared a different set of 40 speakers.

\begin{figure}
\centering
\begin{subfigure}{0.9\linewidth}
    \includegraphics[width=\linewidth]{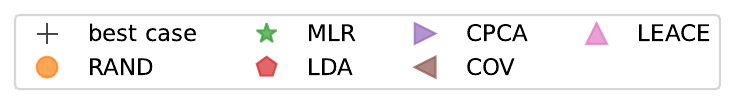} 
\end{subfigure} \\
\vspace{-2pt}
\begin{subfigure}{.49\linewidth}
  \centering
    \includegraphics[width=\linewidth]{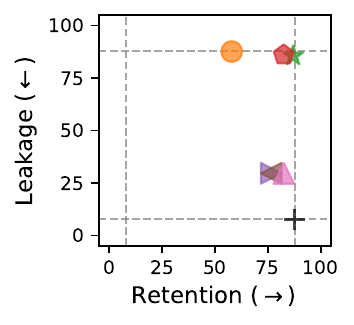} 
  \caption{Phone containment}
  \label{fig:sm_exp:main:sfig1}
\end{subfigure}
\begin{subfigure}{.49\linewidth}
  \centering
  \includegraphics[width=\linewidth]{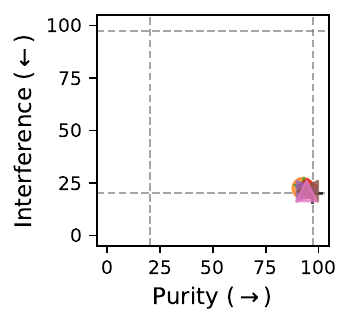} 
  \caption{Phone disentanglement}
  \label{fig:sm_exp:main:sfig2}
\end{subfigure}
\vspace{-2pt}
\begin{subfigure}{.49\linewidth}
  \centering
    \includegraphics[width=\linewidth]{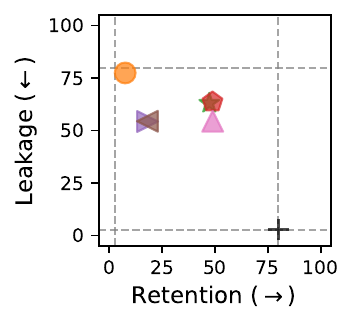} 
  \caption{Speaker containment}
  \label{fig:sm_exp:main:sfig3}
\end{subfigure}
\begin{subfigure}{.49\linewidth}
  \centering
  \includegraphics[width=\linewidth]{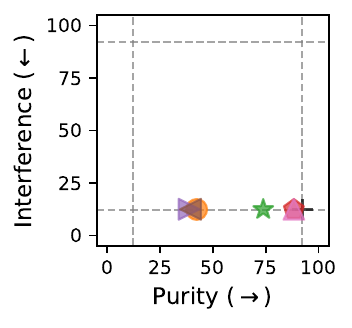} 
  \caption{Speaker disentanglement}
  \label{fig:sm_exp:main:sfig4}
\end{subfigure}
\caption{Containment and disentanglement of phone and speaker subspaces for different subspace estimators. Dashed lines depict best- and worst-case values for each metric.
Numerical results are in Appendix~\ref{appx:exp:sm} Table~\ref{tab:sm_exp:task_space}.
}
\label{fig:sm_exp:main}
\vspace{-10pt}
\end{figure}

\subsection{Phone and speaker subspace estimation}
\label{sec:speech-mainexp}

Fig.~\ref{fig:sm_exp:main} presents results on phone and speaker subspaces for all five concept estimation methods and the random baseline. 
Comparing concept estimators,  the four plots together show that LEACE performs best overall. 
The other variance-based estimators (CPCA, COV) perform similarly on leakage and interference, but often less well on retention and purity.
The probing-based methods (MLR and LDA) have the best retention (near-optimal for phones), but at the cost of very high leakage. 

Despite LEACE's good performance relative to others, we note that it still has substantial leakage (see the left plots), which shows that its no-leakage guarantee does not extend to unseen data. In fact, we found the same issue with generalization when evaluating gender and profession on a fine-tuned BERT model, rather than the pre-trained model from Section~\ref{sec:lm_exp} (see Appendix~\ref{appx:exp:lm}). These results indicate a wider problem with generalization in LEACE,
and reveal important limitations in its use for  concept erasure.

Our results also indicate differences between \emph{how} phonetic and speaker information are represented in HuBERT, which are not obvious either from prior probing studies \cite{pasad_comparative_2023,chrupala_analyzing_2020,martin_probing_2023, Ashihara2024-zo} or from the structural analyses mentioned above, since they did not examine all four measures. Phonetic information seems to be more salient in the representation space than speaker information, as indicated by the relatively high phone retention scores for the random baseline (Fig.~\ref{fig:sm_exp:main:sfig1}, as compared to low retention for RAND in Fig.~\ref{fig:sm_exp:main:sfig3}).
In addition, the best concept estimation methods find phone subspaces with strong containment scores (Fig.~\ref{fig:sm_exp:main:sfig1}), whereas for speaker subspaces, both retention and leakage are considerably sub-optimal for all methods (Fig.~\ref{fig:sm_exp:main:sfig3}). 
The low speaker retention indicates that the 40-dimensional speaker subspace estimated from {\em space-train} does not align well with the optimal 40-dimensional space for distinguishing between the speakers who are included in {\em probe-train} and {\em test}.
We hypothesize that {\em space-train} contains different speakers from the other two sets, making generalization more challenging. 
This hypothesis is further investigated in Section~\ref{sec:more-speakers}.\footnote{In principle, representations from other layers could have different properties. In Appendix~\ref{sec:layerwise}  we present a layerwise analysis using LEACE which confirms prior work indicating that speaker retention is much higher in earlier layers of HuBERT, but also shows that speaker leakage (not measured in prior work) is much higher in those layers. Thus, there is no HuBERT layer where LEACE can find a generalizable well-contained speaker subspace using only 40 training speakers.}

Lastly, we note that the best estimators find speaker and phone subspaces that are highly disentangled from each other  (Figs.~\ref{fig:sm_exp:main:sfig2} and~\ref{fig:sm_exp:main:sfig4}), although purity scores vary for the speaker subspaces. That is, for some methods, including CPCA \cite{Liu2023-sq}, the speaker subspace also contains phonetic information. Nevertheless, our disentanglement results are consistent with those of \citet{Liu2023-sq}, who showed that removing the speaker subspaces did not harm phonetic information.

\subsection{Speaker subspace generalization}
\label{sec:more-speakers}

One strength of our framework is that it makes explicit the step of subspace estimation and the step of evaluating the four measures.
In other words, the framework includes two types of generalization: the generalization of subspace estimation and the generalization of the probes for the measurements.
In this section, we focus on the generalization of subspace estimation, in particular the speaker subspaces estimated in Section~\ref{sec:speech-mainexp}.

The speaker subspaces estimated on space-train did not generalize well, indicating likely overfitting to the particular set of 40 speakers in space-train.
To test this hypothesis, we compared three data splits for estimating speaker subspaces using LEACE: space-train (standard setup), probe-train (overfitting setup), and LibriSpeech's train-clean-460, containing 1172 speakers (scaling setup). We considered probe-train as the overfitting setup because the same data (thus, same set of speakers) is used to both estimate speaker subspaces and train the probing classifiers.

\begin{figure}
\centering
\begin{subfigure}{1\linewidth}
  \centering
    \includegraphics[width=\linewidth]{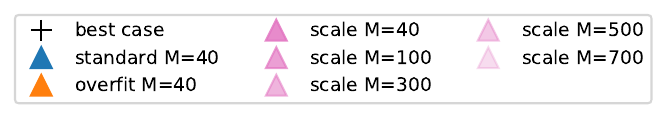} 
\end{subfigure} \\
\vspace{-3pt}
\begin{subfigure}{.49\linewidth}
  \centering
    \includegraphics[width=\linewidth]{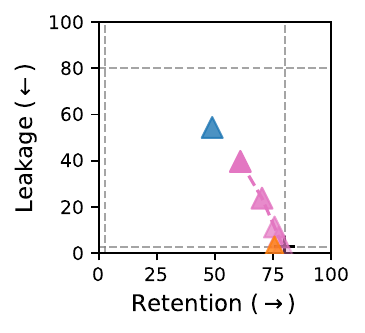} 
  \caption{Speaker containment}
  \label{fig:sm_exp:spk_gen:sfig1}
\end{subfigure}
\begin{subfigure}{.49\linewidth}
  \centering
  \includegraphics[width=\linewidth]{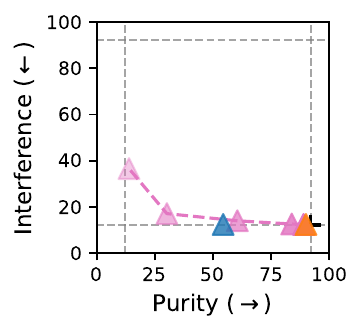} 
  \caption{Speaker disentanglement}
  \label{fig:sm_exp:spk_gen:sfig2}
\end{subfigure}
\caption{Generalization of speaker subspaces to unseen data and different dimensionality. Standard and overfit setups have 40 speakers, while the scale setup has 1172 speakers. $M$ is the dimensionality of the subspace.
For numerical results, see Appendix~\ref{appx:exp:sm}, Table \ref{tab:sm_exp:spk_gen}.
}
\vspace{0em}
\label{fig:sm_exp:spk_gen}
\end{figure}

We found that the overfitting setup yields near-optimal scores for both containment and disentanglement (Fig.~\ref{fig:sm_exp:spk_gen}, orange markers),
substantially outperforming the standard setup (blue markers). In particular, leakage is optimal, aligning with the proof shown by prior work \cite{Belrose2023-ny}.

Fig.~\ref{fig:sm_exp:spk_gen} also shows that 
if we keep the dimensionality of the speaker space at $M=40$,
scaling up the number of speakers does improve both containment and disentanglement (darkest pink marker). 
However, the improvement in containment scores is relatively small, suggesting that even with a large number of speakers, it may be difficult to find a low-dimensional subspace that generalizes well. 

To further explore this issue, we increased the dimension of the speaker subspace, ranging up to $M=700$.  Fig.~\ref{fig:sm_exp:spk_gen:sfig1} shows that increasing $M$ steadily improves containment.
However, Fig.~\ref{fig:sm_exp:spk_gen:sfig2} shows that disentanglement degrades as $M$ increases, a trade-off analogous to those found in causal intervention spaces \cite{huang_ravel_2024}. We found a steady drop in purity, implying more phone information leaks into speaker subspaces at higher dimensions. Interestingly, interference remains low up to $M=500$, indicating that phone information is highly robust to the erasure of speaker subspaces.

Overall, our results suggest that estimating a robust concept space for speaker identity requires a large number of speakers as well as a large number of dimensions.
Moreover, although there is a high degree of disentanglement (orthogonality) between the phone and speaker spaces, linear speaker information cannot be fully removed without impacting phone recognition.

\section{Discussion and Conclusion}
In this work, we introduced a novel framework to study concept subspace structures by evaluating containment and disentanglement.
By providing new insights on representations from both BERT and HuBERT, we demonstrated the framework's utility across different models and modalities, and highlighted the importance of the concept estimator in analyses of concept representation. 
Our experiments indicated the LEACE estimator as the most effective method among five tested estimators for identifying concept subspaces with strong containment and disentanglement properties, although even LEACE struggled in some cases to generalize beyond its training data.

Our framework has several extensions to be addressed in the future work.
For example, we assumed discrete concepts here, but the framework could be extended to handle continuous concepts, such as pitch or energy in speech  representations, by using regression models.  We also only examined pairwise disentanglement, but multi-way disentanglement could potentially be addressed by aggregating disentanglement scores across multiple concepts.

Our framework also opens the way for investigating additional questions.
While we mainly investigated concept subspaces whose dimensionality is equal to the number of concept classes, \citet{Hernandez2021-nv} showed that some linguistic concepts, with fixed label sets, are decodeable from a subspace with dimensionality even lower than the number of classes. 
However, our results suggest that when the label set is not fixed 
(i.e., the labels in space-train and probe-train/test are disjoint) 
estimating a low-dimensional well-contained concept subspace becomes significantly harder (Section \ref{sec:more-speakers}). This challenge likely extends to any concepts with unbounded classes, such as facial identities and languages spoken in the world, where any concept space must be estimated from just a sample of the classes. Studying which concepts are represented with fewer dimensions than labels, and why, is an interesting future direction.

To sum up, we believe that by formalizing disparate notions of concept subspaces under a single framework, highlighting the role of the concept estimator, and establishing a comprehensive set of measures, our work will help to inspire and solidify future work on concept representation in neural models.

\bibliography{paperpile_local.bib, references_sgwater_local.bib}

\newpage
\appendix

\section{Limitations} 
\paragraph{Concept pairwise evaluation} Our current framework is defined for evaluating disentanglement using concept pairs, rather than capturing interactions among multiple concepts. A possible extension is to aggregate results across purity and interference scores measured against different concepts.

\paragraph{Cross-model comparability} Our metrics are not directly comparable across models, since they depend on the ambient space performance of each model. This baseline must be taken into account when interpreting results.

\paragraph{Cross-task comparability} Our metrics are not straightforwardly interpretable when comparing across tasks. Metrics are inherently easier for binary classification and harder for tasks with many classes. Although majority-prediction baselines can be used for calibration, they do not fully resolve these scale differences.

\paragraph{Dependence on representation quality} Concept subspaces with poor retention often exhibit good leakage, purity, or interference scores. However, this is largely a by-product of weak representations. Our method is not very meaningful if the concept accuracy on the ambient space is already low.

\paragraph{Linearity} Our evaluation assesses concept subspaces using only linear probes, while it is also possible that concept subspaces contains non-linear structures. However, the framework itself is not limited to linear probes, and extending the evaluation to non-linear probes could be possible in future work.

\paragraph{Empirical limitations} Although we evaluated several different estimators on two different modalities of data and four different concepts, our empirical evaluation is necessarily limited. Since our conclusions about estimators were based on two completely different models and modalities, we believe these are likely to generalize to other settings. However, the conclusions about the specific concepts studied here may not hold for representations extracted from other models or datasets, and could be explored in future work.

\section{Example studies using different concept estimation methods}
\label{appx:estimation}

Table \ref{tab:appx:estimator_cmpr} compares concept estimators and how they were used in previous works, describing the goals of each study and the models and concepts that were investigated. This is not an exhaustive list, but provides some indication of the types of questions and studies.

\begin{table*}[]
    \centering
    \begin{tabular}{|p{3cm}|p{12cm}|}
    \hline
       {\bf Estimator} & {\bf Example studies}\\
       \hline
       MLR  & \citet{Hernandez2021-nv} investigated if language models (ELMo and BERT) encode part-of-speech, dependency label, and dependency edge concepts in low-dimensional spaces, and argued that the geometric structure of those subspaces is organized hierarchically. 
       \citet{Ravfogel2020-jj} de-biased BERT representations by applying nullspace projection on the gender subspaces.
       \\ 
       \hline
       LDA & \citet{Chang2022-ky} studied the structure of language and token position subspaces of a multilingual language model (XLM-R). \citet{Park2024-ox} estimated directions that causally change predictions of Gemma and LLaMA-3.  \\
       \hline CPCA & \citet{Liu2023-sq} explored relationships between phone and speaker subspaces in self-supervised predictive coding speech models (APC and CPC). \citet{Mohamed2024-xe} extended their work to other self-supervised models (wav2vec2.0, HuBERT, WavLM).\\
       \hline COV  & \citet{Shao2023-nn} erased gender and demographic information to de-bias BERT.\\
       \hline LEACE & \citet{Belrose2023-ny} proposed LEACE to de-bias BERT. \citet{Krishnan2024-nx} erased gender in Wav2Vec and HuBERT speech models. \citet{saillenfest_nonlinear_2025} used LEACE as a sub-function of a nonlinear erasure method, proposing to de-bias BERT and DeepMoji.  \\
    \hline
    \end{tabular}
    \caption{Example studies using each of the concept estimators we explore. We  focus on studies that explicitly use these  methods to identify concept subspaces (rather than just for, e.g., classification).}
    \label{tab:appx:estimator_cmpr}
\end{table*}

\section{Additional results for gender and profession case study}
\label{appx:exp:lm}
In this section, we provide results for all estimators in the gender and profession subspace experiments in Section~\ref{sec:lm_exp}, and analogous results for representations extracted from BERT after fine-tuning on the profession classification task (using the model from \citet{Ravfogel2022-cg}).

\begin{figure}[]
\centering
\begin{subfigure}{0.8\linewidth}
    \includegraphics[width=\linewidth]{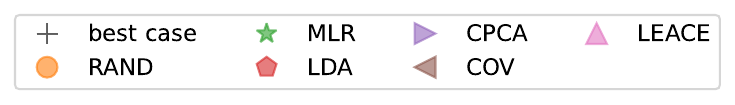}  
\end{subfigure} \\
\vspace{-4pt}
\begin{subfigure}{.49\linewidth}
    \includegraphics[width=\linewidth]{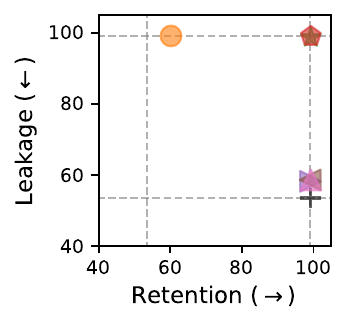} 
  \caption{Gender containment}
  \label{fig:full_text_exp:sfig1}
\end{subfigure}
\begin{subfigure}{.49\linewidth}
  \includegraphics[width=\linewidth]{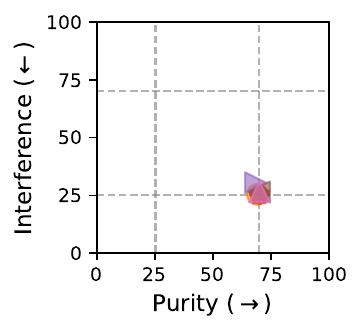} 
  \caption{Gender disentanglement}
  \label{fig:full_text_exp:sfig2}
\end{subfigure}
\begin{subfigure}{.49\linewidth}
    \includegraphics[width=\linewidth]{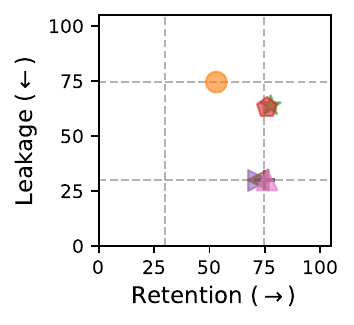} 
  \caption{Prof. containment}
  \label{fig:full_text_exp:sfig3}
\end{subfigure}
\begin{subfigure}{.49\linewidth}
  \includegraphics[width=\linewidth]{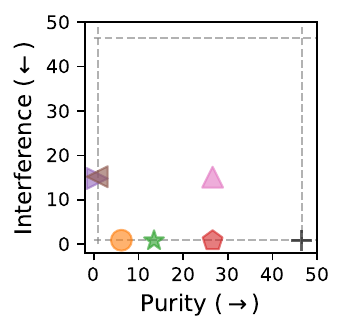} 
  \caption{Prof. disentanglement}
  \label{fig:full_text_exp:sfig4}
\end{subfigure}
\caption{
Containment and disentanglement of gender and profession (Prof.) subspaces for all estimators. The representations are the CLS tokens of pretrained BERT. For numerical results, see Table \ref{tab:lm_exp:main_frozen_space}.
}
\label{fig:full_text_exp:main_fz}
\vspace{-8pt}
\end{figure}

As an extension of Section~\ref{sec:lm_exp}, we present a comprehensive comparison for all concept estimators in Fig.~\ref{fig:full_text_exp:main_fz}. This analysis shows that estimators within the same group find concept subspaces with similar properties. In particular, probing-based estimators (MLR and LDA) find subspaces with high leakage and low interference, whereas the variance-based estimators (CPCA, COV, and LEACE) tend to find subspaces that have lower leakage but higher interference. The detailed numerical results can be found in Table~\ref{tab:lm_exp:main_frozen_space}. Note that, for the RAND estimator, we averaged the results over five random subspaces to reduce the effects of random seeds.

\begin{figure}[]
\centering
\begin{subfigure}{0.8\linewidth}
    \includegraphics[width=\linewidth]{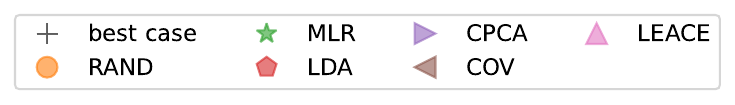}  
\end{subfigure} \\
\vspace{-4pt}
\begin{subfigure}{.49\linewidth}
    \includegraphics[width=\linewidth]{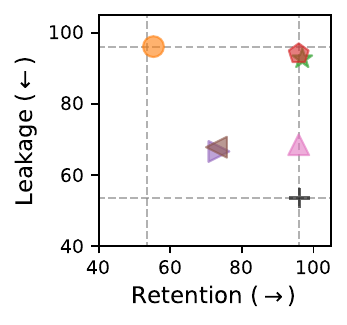} 
  \caption{Gender containment}
  \label{fig:full_text_exp:main_ft:sfig1}
\end{subfigure}
\begin{subfigure}{.49\linewidth}
  \includegraphics[width=\linewidth]{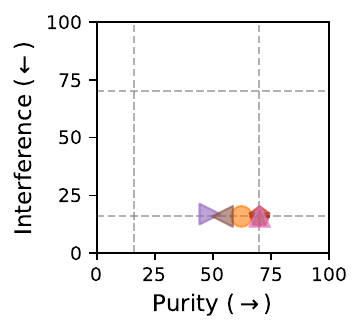} 
  \caption{Gender disentanglement}
  \label{fig:full_text_exp:main_ft:sfig2}
\end{subfigure}
\begin{subfigure}{.49\linewidth}
    \includegraphics[width=\linewidth]{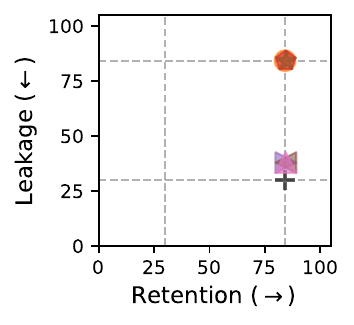} 
  \caption{Prof. containment}
  \label{fig:full_text_exp:main_ft:sfig3}
\end{subfigure}
\begin{subfigure}{.49\linewidth}
  \includegraphics[width=\linewidth]{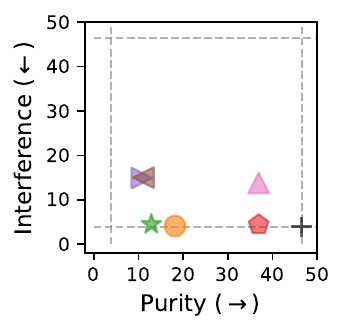} 
  \caption{Prof. disentanglement}
  \label{fig:full_text_exp:main_ft:sfig4}
\end{subfigure}
\caption{
Containment and disentanglement of gender and profession (Prof.) subspaces for all estimators. The representations are the CLS tokens of a BERT model finetuned on profession classification. For numerical results, see Table~\ref{tab:lm_exp:main_finetuned_space}.
}
\label{fig:full_text_exp:main_ft}
\vspace{-8pt}
\end{figure}

For the fine-tuned representations, we observe that profession information becomes more salient, as demonstrated explicitly by the improved retention of the RAND-estimated profession subspace (Fig.~\ref{fig:full_text_exp:main_ft:sfig3}) and implicitly by the increased profession purity across all estimators (Fig.~\ref{fig:full_text_exp:main_ft:sfig4}). In contrast, the reduced retention of the RAND-estimated gender subspace (Fig.~\ref{fig:full_text_exp:main_ft:sfig1}) and the overall decrease in gender purity (Fig.~\ref{fig:full_text_exp:main_ft:sfig2}) indicate that gender information is suppressed. Moreover, leakage for both gender and profession subspaces increases noticeably, suggesting that fine-tuning impact the linear subspace structure of these concept representations, potentially increasing redundancy. Note that all reported values are averaged over five model instances fine-tuned with different random seeds for each estimator. Numerical details are provided in Table~\ref{tab:lm_exp:main_finetuned_space}.

\begin{table}[]
\centering
\scalebox{0.9}{
    \begin{tabular}{llllll}
    \hline
    Concept & Estimator & Ret.$\uparrow$ &Leak.$\downarrow$ & Pur.$\uparrow$ &Int.$\downarrow$ \\
    \hline
    gender & worst case& 53.5 & 99.1 & 25.3 & 70.0 \\
    gender & RAND  & 60.3 & 99.1 & 70.0 & 25.3 \\ 
    \hline
    gender & MLR    & 99.3 & 98.7 & 70.0 & 25.5 \\
    gender & LDA   & 99.3 & 99.0 & 70.0 & 25.3 \\
    gender & CPCA   & 99.2 & 58.2 & 68.3 & 30.5 \\
    gender & COV   & 99.1 & 58.6 & 70.0 & 26.5 \\
    gender & LEACE & 99.3 & 58.6 & 70.0 & 26.5 \\
    \hline
    gender & best case& 99.1 & 53.5 & 70.0 & 25.3 \\
    \hline \hline
    prof   & worst case& 30.0 & 74.7 & 0.9  & 46.5 \\
    prof   & RAND  & 53.4 & 74.3 & 4.6  & 0.9  \\ 
    \hline
    prof   & MLR    & 77.7 & 63.8 & 13.5 & 0.8  \\
    prof   & LDA   & 76.0 & 63.0 & 26.6 & 0.8  \\
    prof   & CPCA   & 72.2 & 29.8 & 0.8  & 14.8 \\
    prof   & COV   & 72.0 & 30.0 & 0.8  & 15.2 \\
    prof   & LEACE & 76.0 & 30.0 & 26.6 & 15.1 \\
    \hline
    prof   & best case& 74.7 & 30.0 & 46.5 & 0.9  \\
    \hline
    \end{tabular}
}
\caption{Numerical results for containment and disentanglement of the gender and profession subspaces in pretrained BERT representations. For visualization, see Fig.~\ref{fig:full_text_exp:main_fz}.}
\label{tab:lm_exp:main_frozen_space}.
\end{table}

\begin{table}[]
\centering
\scalebox{0.9}{
    \begin{tabular}{llllll}
    \hline
    Concept & Estimator & Ret.$\uparrow$ &Leak.$\downarrow$ & Pur.$\uparrow$ &Int.$\downarrow$ \\
    \hline
    gender & worst case & 53.5 & 96.1 & 15.9 & 70.0 \\
    gender & RAND  & 55.3 & 96.1 & 62.3 & 15.9 \\
    \hline
    gender & MLR    & 96.9 & 92.7 & 69.9 & 15.9 \\
    gender & LDA   & 95.9 & 94.1 & 70.1 & 15.8 \\
    gender & CPCA   & 73.6 & 66.6 & 48.7 & 16.9 \\
    gender & COV   & 72.9 & 67.8 & 54.1 & 15.9 \\
    gender & LEACE & 95.9 & 68.5 & 70.1 & 15.9 \\
    \hline 
    gender & best case  & 96.1 & 53.5 & 70.0 & 15.9 \\
    \hline \hline
    prof   & worst case & 30.0 & 84.1 & 3.9  & 46.5 \\
    prof   & RAND  & 84.1 & 84.2 & 18.2 & 4.1  \\
    \hline
    prof   & MLR    & 84.7 & 84.1 & 12.9 & 4.5  \\
    prof   & LDA   & 84.4 & 84.2 & 36.9 & 4.4  \\
    prof   & CPCA   & 84.5 & 37.9 & 10.8 & 14.9 \\
    prof   & COV   & 84.5 & 37.9 & 11.1 & 15.0 \\
    prof   & LEACE & 84.4 & 38.0 & 36.9 & 13.8 \\
    \hline
    prof   & best case  & 84.1 & 30.0 & 46.5 & 3.9 \\
    \hline
    \end{tabular}
}
\caption{Numerical results for containment and disentanglement of the gender and profession subspaces in BERT representations finetuned on profession classification. For visualization, see Fig.~\ref{fig:full_text_exp:main_ft}.}
\label{tab:lm_exp:main_finetuned_space}
\end{table}

\section{Additional results for phone and speaker case study}
\label{appx:exp:sm}

This section presents numerical results for the phone and speaker subspace comparison described in Section~\ref{sec:sm_exp}, along with an additional layer-wise analysis of phone and speaker subspace properties across all HuBERT layers.

We present the numerical results used to generate Figure~\ref{fig:sm_exp:main} in Table~\ref{tab:sm_exp:task_space}. To reduce the influence of random seeds, we report an average computed over five random subspaces for the RAND-estimated subspace.

Table~\ref{tab:sm_exp:spk_gen} reports the numerical results corresponding to Figure~\ref{fig:sm_exp:spk_gen} and provides additional results for subspaces with dimensionalities of 200, 400, 600, and 768. Note that the subspace with M=768 spans the entire ambient space. Consequently, the retention and leakage scores reach their best-case values. In contrast, purity and interference attain their worst-case values, as no complementary subspace remains (i.e., the kernel dimensionality is zero).

\begin{table}[]
\centering
\scalebox{0.9}{
    \begin{tabular}{llllll}
    \hline
    Concept & Estimator & Ret.$\uparrow$ &Leak.$\downarrow$ & Pur.$\uparrow$ &Int.$\downarrow$ \\
    \hline
    phones   & worst case     & 7.8  & 87.8  & 20.1 & 97.3 \\
    phones   & RAND      & 57.9 & 87.6  & 92.5 & 22.5 \\ 
    \hline
    phones   & MLR        & 87.0 & 85.6  & 93.1 & 22.6 \\
    phones   & LDA       & 82.5 & 86.0  & 94.6 & 22.0 \\
    phones   & CPCA       & 76.6 & 29.6  & 94.6 & 21.1 \\
    phones   & COV       & 76.6 & 29.6  & 94.8 & 21.1 \\
    phones   & LEACE     & 82.6 & 29.6  & 94.3 & 21.2 \\
    \hline
    phone   & best case      & 87.8 & 7.8   & 97.3 & 20.1 \\
    \hline \hline 
    speakers     & worst case     & 2.7  & 79.9  & 12.2 & 92.2 \\
    speakers     & RAND      & 7.5  & 77.4  & 42.3 & 12.4 \\ 
    \hline
    speakers     & MLR        & 47.3 & 63.0  & 73.8 & 12.4 \\
    speakers     & LDA       & 48.8 & 63.6  & 88.3 & 12.4 \\
    speakers     & CPCA       & 18.1 & 54.3  & 38.7 & 12.5 \\
    speakers     & COV       & 18.0 & 54.4  & 39.4 & 12.5 \\
    speakers     & LEACE     & 48.9 & 54.4  & 88.3 & 12.5 \\
    \hline
    speakers     & best case      & 79.9 & 2.7   & 92.2 & 12.2 \\
    \hline
    \end{tabular}
}
\caption{Phonetic and speaker subspaces for speech models, for visualization see Section~\ref{sec:speech-mainexp} Fig.~\ref{fig:sm_exp:main}.}
\label{tab:sm_exp:task_space}
\end{table}

\begin{table}[]
\centering
\scalebox{0.9}{
    \begin{tabular}{lllllll}
    \hline
    Trainset    & Dim. &  Ret.$\uparrow$ &Leak.$\downarrow$ & Pur.$\uparrow$ &Int.$\downarrow$ \\
    \hline
    worst case           & -         & 2.7  & 79.9 & 12.2 & 92.2 \\
    \hline
    standard & M=40  & 48.8 & 75.6 & 54.4 & 90.0 \\
    overfit & M=40  & 54.5 & 2.7  & 12.4 & 12.4 \\
    \hline
    scale    & M=40  & 60.9 & 39.7 & 89.0 & 12.4 \\
    scale    & M=100 & 70.2 & 23.8 & 83.9 & 12.6 \\
    scale    & M=200 & 73.8 & 15.4 & 73.5 & 13.2 \\
    scale    & M=300 & 75.6 & 11.3 & 60.5 & 14.0 \\
    scale    & M=400 & 76.7 & 8.7  & 44.9 & 15.2 \\
    scale    & M=500 & 77.4 & 6.9  & 30.2 & 17.1 \\
    scale    & M=600 & 78.3 & 5.3  & 19.0 & 21.3 \\
    scale    & M=700 & 79.0 & 3.7  & 13.9 & 36.6 \\
    scale    & M=768 & 80.0 & 2.7  & 12.2 & 92.1 \\
    \hline
    best case           & -          & 80.0 & 2.7  & 92.1 & 12.2 \\
    \hline
    \end{tabular}
}
\caption{Generalization of speaker subspaces to unseen speakers, estimated using LEACE with different dimensionality (M). For visualization, see Section~\ref{sec:more-speakers} Fig.~\ref{fig:sm_exp:spk_gen}.}
\label{tab:sm_exp:spk_gen}
\end{table}

\begin{figure*}[]
\centering
\begin{subfigure}{1\textwidth}
    \includegraphics[width=\linewidth]{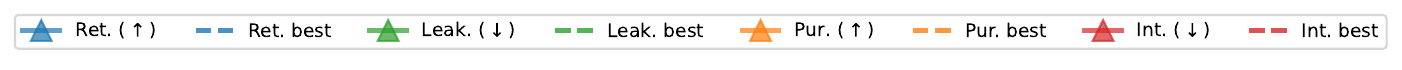} 
\end{subfigure} \\
\vspace{-7pt}
\begin{subfigure}{.24\textwidth}
  \centering
    \includegraphics[width=\linewidth]{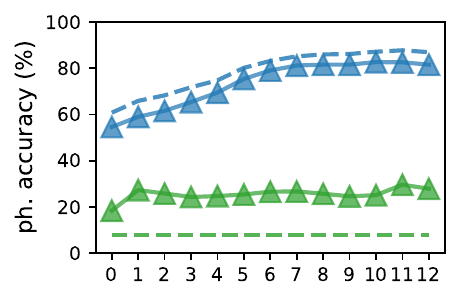} 
  \caption{Phone containment}
  \label{fig:sm_exp:layerwise_cmpr:sfig1}
\end{subfigure}
\begin{subfigure}{.24\textwidth}
  \centering
  \includegraphics[width=\linewidth]{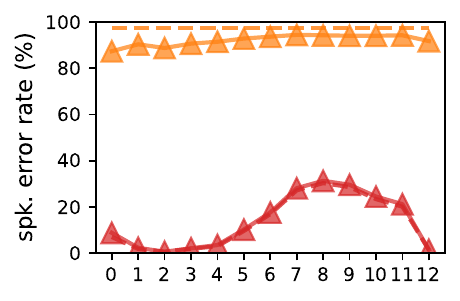} 
  \caption{Phone disentanglement}
  \label{fig:sm_exp:layerwise_cmpr:sfig2}
\end{subfigure}
\begin{subfigure}{.24\textwidth}
  \centering
    \includegraphics[width=\linewidth]{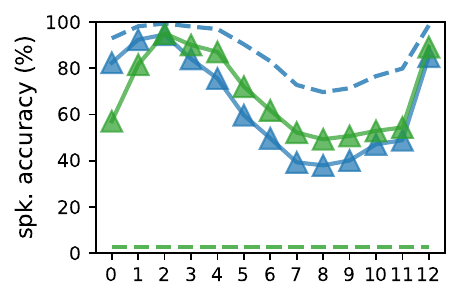} 
  \caption{Speaker containment}
  \label{fig:sm_exp:layerwise_cmpr:sfig3}
\end{subfigure}
\begin{subfigure}{.24\textwidth}
  \centering
  \includegraphics[width=\linewidth]{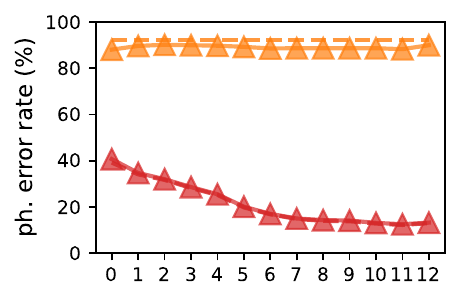} 
  \caption{Speaker disentanglement}
  \label{fig:sm_exp:layerwise_cmpr:sfig4}
\end{subfigure}
\caption{Containment and disentanglement of phone and speaker subspaces estimated by LEACE across HuBERT layers. The x-axis indicates layer depth, where 0 corresponds to the output of the CNN, and 1–12 refer the subsequent Transformer layers.}
\label{fig:sm_exp:layerwise_cmpr}
\vspace{-8pt}
\end{figure*}

\subsection{Layer-wise analysis for phone and speaker concepts} 
\label{sec:layerwise}

As we primarily use representations from layer 11 of the HuBERT model in our experiments, it is natural to ask whether the observations generalize to other layers. This section explores representations all HuBERT layers and finds that the distinct properties of the phone and speaker subspaces hold consistently throughout the model.

Fig.~\ref{fig:sm_exp:layerwise_cmpr:sfig1} shows that the phone subspaces reach near-optimal retention in every layer, where the deeper layers have higher accuracies (consistent with probing results from \citealp{pasad_comparative_2023, Mohamed2024-xe}). Meanwhile, leakage remains largely stable. Together, they indicate strong phone containment across layers of HuBERT.

In contrast, Fig.~\ref{fig:sm_exp:layerwise_cmpr:sfig3} shows that retention and leakage for speaker subspaces are generally suboptimal, positioned away from their best cases. Early layers exhibit better retention, while middle layers exhibits better leakage.
Moreover, retention and leakage vary sharply together, although they should ideally diverge (as observed in Fig.~\ref{fig:sm_exp:layerwise_cmpr:sfig1} for phone containment). This results indicates that no well-contained speaker subspace is found in any layer.

For completeness, we also examined disentanglement across layers (Figs.~\ref{fig:sm_exp:layerwise_cmpr:sfig2} and~\ref{fig:sm_exp:layerwise_cmpr:sfig4}). In line with the results from Section~\ref{sec:speech-mainexp}, we found that interference was consistently optimal across all layers for both types of subspace, indicating that removing one of the subspaces would not damage the other. Meanwhile purity was close to optimal except a slight drop for phones in the early layers. 
These show that HuBERT disentangles phone and speaker subspaces across layers.

\end{document}